\documentclass[runningheads]{llncs}

\usepackage{eccv}

\usepackage{eccvabbrv}
\usepackage{array}
\usepackage{utfsym}
\usepackage{multirow}
\usepackage{graphicx}
\usepackage{booktabs}
\usepackage{dsfont}

\usepackage[accsupp]{axessibility}  

\usepackage{hyperref}

\usepackage{orcidlink}

\begin{document}

\title{Continuity Preserving Online CenterLine Graph Learning} 

\titlerunning{CGNet}

\author{Yunhui Han \and
Kun Yu \and
Zhiwei Li$^\dagger$}

\authorrunning{Y. Han et al.}

\institute{Xiaomi EV, Beijing, China \\
\email{\{hanyunhui1,yukun,lizhiwei7\}@xiaomi.com}\\
$^\dagger$Corresponding Author}

\maketitle

\begin{abstract}
Lane topology, which is usually modeled by a centerline graph, is essential for high-level autonomous driving. For a high-quality graph, both topology connectivity and spatial continuity of centerline segments are critical. However, most of existing approaches pay more attention to connectivity while neglect the continuity. Such kind of centerline graph usually cause problem to planning of autonomous driving. To overcome this problem, we present an end-to-end network, CGNet, with three key modules: 1) Junction Aware Query Enhancement module, which provides positional prior to accurately predict junction points; 2) Bézier Space Connection module, which enforces continuity constraints on any two topologically connected segments in a Bézier space; 3) Iterative Topology Refinement module, which is a graph-based network with memory to iteratively refine the predicted topological connectivity. CGNet achieves state-of-the-art performance on both nuScenes and Argoverse2 datasets. Our code is available at \url{https://github.com/XiaoMi/CGNet}.
\end{abstract}

\section{Introduction}
\label{sec:intro}
High-definition(HD) map is a base of many autonomous driving components, such as motion planning and  forecasting\cite{gao2020vectornet,jiang2023vad, zhao2021tnt}. However, constructing of HD map demands a complex pipeline and manual annotation, which is costly, labor-intensive and lack of freshness. Recently, online map reconstruction by visual perception has attracted lots of attention. Many works \cite{shin2023instagram,ding2023pivotnet,qiao2023end} like VectormapNet\cite{liu2023vectormapnet} and MapTR\cite{liao2022maptr} directly obtain vectorized map in an end-to-end manner and achieve promising results. However, these methods usually focus on detecting visible road markers (e.g. lane divider, road boundary and pedestrian crossing), but pay little attention to lane topology.

\begin{figure}[tp]
    \centering
    \includegraphics[width=0.7\linewidth]{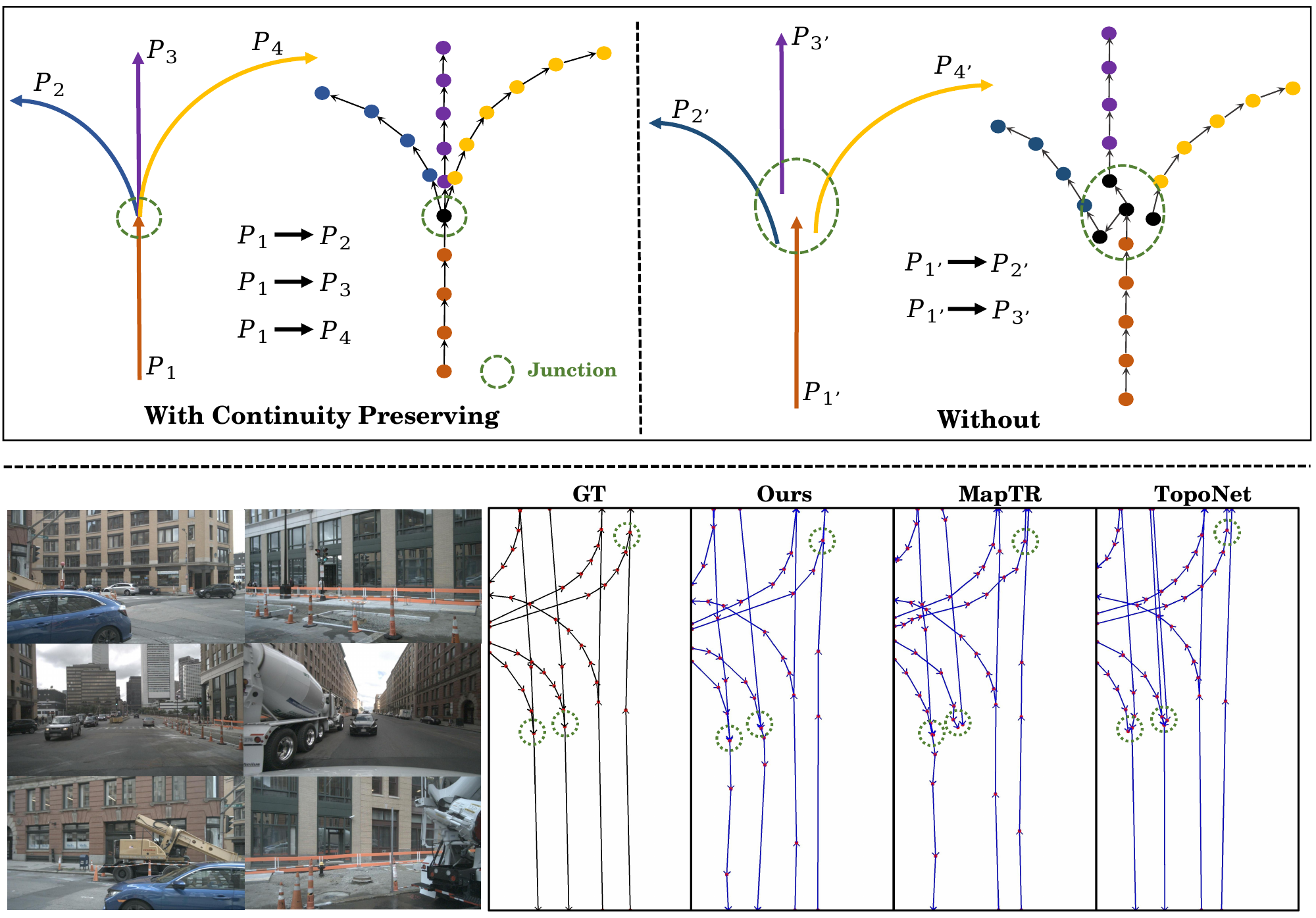}
    \caption{ The motivation. Top: A toy example which illustrates the centerline graph and the impact of overlooking the continuity. Bottom: Comparison with MapTR and TopoNet. They predicts inaccurate position of junction points and wrong topology, all leading to the discontinuous path. Our CGNet obtain the continuous path.}
    \label{fig:teaser}
\end{figure}

 In literature, lane topology is usually modeled by a graph of lane segments \cite{can2021structured}, which is often termed as centerline graph as shown in Fig.~\ref{fig:teaser}. A centerline graph has two key attributes: lane segment position and topology. Most of existing methods can predict segment position properly, yet still struggle in recovering topology. The STSU series of works\cite{can2021structured,can2022topology,can2023improving} and TopoNet\cite{li2023topology} dedicate to modelling centerline graph. The main issue of these methods is that centerline segments are treated as independent entities without explicitly enforcing \emph{continuity} constraints. In real-world driving scenarios, autonomous vehicles should drive along a smooth path instead of spatially discontinuous path segments. A toy example in Fig.~\ref{fig:teaser} top illustrates the problem of overlooking the continuity. Even if the topological connectivity of lane segments is correct, the graph cannot be used in autonomous driving due to spatial discontinuity.
 
 To preserve continuity, LaneGAP\cite{liao2023lane} proposes to detect complete lanes (with overlapped segments) instead of non-overlapped lane segments. A post-processing is adopted to merge individually detected points/lanes to obtain a topologically sound lane graph. These steps usually depend on manually tuned thresholds. RNTR\cite{lu2023translating} establishes a bijection from lane graph to lane sequence and proposes a autoregressive sequence-to-sequence model, which satisfies the continuity requirement but is computationally expensive.

In this work, we present Continuity Preserving Online CenterLine Graph Learning Network(CGNet), which follow the scheme of modeling centerline graph in a segmented manner\cite{can2021structured} and focus on preserving the continuity. Fig.~\ref{fig:overview} shows the overall structure of our network, which bases on a vectorized online map construction method\cite{liao2022maptr}. Intuitively, three factors are helpful for improving continuity: accurate positions of junction points, spatial smoothness of logically connected centerline segments and the correct connectivity. Three modules, which specially designed from local to global perspective, are used to meet these conditions: 1) \emph{Junction Aware Query Enhancement Module(JAQ)}. It extracts junction points feature from BEV feature and makes the feature interact with lane queries to provide positional prior for queries, which enhance the positional perception ability of queries. The continuity can be better ensured when junction points are predicted more accurately. 2) \emph{Bézier Space Connection Module(BSC)}. It projects embeddings of any two connected centerline segments into a Bézier space and predict control points in a compact form. The lane embeddings are constrained on connectivity in an implicit way, which improves the continuity and smoothness of centerline. 3) \emph{Iterative Topology Refinement Module(ITR)}. It predicts connectivity of all centerline segments in an iterative manner by combining a GCN and a GRU. By capturing information from previous outputs, the module is able to improve connectivity accuracy incrementally.

In summary, our main contributions are: 
\begin{itemize}
    \item We present \emph{CGNet}, an end-to-end network for online centerline graph understanding, which explicitly handles the discontinuity problem.
    \item We elaborately design \emph{Junction Aware Query Enhancement module}, \emph{Bézier Space Connection module} and \emph{Iterative Topology Refinement module} to improve the continuity from different perspectives.
    \item The proposed CGNet achieves a new state-of-the-art performance on the challenging nuScenes and Argoverse2 dataset. 
\end{itemize}

\section{Related Work}
\textbf{Lane Detection.} Lane detection is a fundamental work for autonomous driving. Previous works utilize various forms to represent lanes. In 2D lane detection, most works\cite{li2019line, tabelini2021keep,ko2021key} like SCNN\cite{pan2018spatial} model the lane as a set of points in image frame. LSTR\cite{liu2021end} adopts a cubic curve to approximate a single lane line and BézierLaneNet\cite{feng2022rethinking} uses Bézier curve to model the lane. In 3D lane detection, most works\cite{chen2022persformer, yao2023sparse, luo2023latr} like 
LaneNet\cite{garnett20193d} directly uses the real-world 3D road coordinates to represent lanes. Curveformer\cite{bai2023curveformer} represents lane by the curve parameters in 3D space. In general, representing lane by a parametric curve can better ensure the consistency and smoothness, while representing lane as a set of points is more flexible and fitting for ground-truth. These methods provide good insights into lane representation, but they usually detect lanes in a single view, which is insufficient for high-level driving.

\textbf{Online Map Reconstruction.} With the popularity of BEV (Bird's Eye View) \cite{li2022bevformer,philion2020lift} representation, early works\cite{peng2023bevsegformer,zhou2022cross,pan2020cross,roddick2020predicting,lu2019monocular,hu2021fiery,li2022hdmapnet} formulate map reconstruction as a segmentation problem, which predict rasterized map from BEV perspective. However, for use of downstream tasks, we need a complicated and time-consuming post-processing to vectorize a rasterized map. To alleviate this issue, VectormapNet\cite{liu2023vectormapnet} utilizes auto-regressive decoder to directly predict vectorized outputs without requiring post-processing. InstaGraM\cite{shin2023instagram} proposes an efficient network which predicts a set of vertices of the road elements and associates these vertices to get a map. MapTR\cite{liao2022maptr} treats online vectorized map reconstruction as a parallel regression problem, all instances and all points of instance are predicted simultaneously. BeMapNet\cite{qiao2023end} treats the problem in 
a similar way like MapTR, but models the line using Bézier curve instead of polyline. PivotNet\cite{ding2023pivotnet} proposes to represent lanes by dynamic point sequences rather than fixed number of points. Unfortunately, these methods only focus on lane detection but are lack the ability to model topology.

\textbf{Lane Graph Learning.} Extracting lane graph from aerial and satellite imagery\cite{he2022lane,buchner2023learning,bastani2018roadtracer,he2020sat2graph,li2019topological} has been studied for many years. However, aerial images have a lower resolution and roads in them are often occluded by other objects like trees. Therefore, these approaches are only able to extract road-level structure, and cannot be used for fine-grained autonomous driving. Another line of work is lidar-based\cite{homayounfar2018hierarchical,homayounfar2019dagmapper} lane graph construction. However, lidar could not recognize some line properties like colors. CenterLineDet\cite{xu2022centerlinedet} designs
a centerline graph detection model with both camera and lidar, but works in an offline manner. STSU\cite{can2021structured} is a pioneering work, which constructs lane graph online using centerline segments from a single onboard camera. Based on STSU, TPLR\cite{can2022topology} introduces the concept of minimal cycles to better capture topology and OLC\cite{can2023improving} jointly detect objects and produce lane graph. However, they work in a limited-FOV perception range and neglect the continuity of centerline segments. TopoNet\cite{li2023topology} designs a graph-based network to promote the connection relationship prediction among centerline segments using surround-view cameras, but also without explicit constraints on continuity. LaneGAP\cite{liao2023lane} proposes to connect all centerline segments into a complete path, performs path detection and finally merges the overlapped vertices of different paths for preserving continuity. However, it is challenging to determine the distance threshold for the merging in post-processing. RNTR\cite{lu2023translating} establishes a bijection lane graph of lane sequence and proposes an autoregressive sequence-to-sequence model, but it is computationally expensive. Inspired by prior works, the proposed CGNet models lane graph by segmenting lanes to non-overlapped segments, and designs three novel modules to address the spatially discontinuous problem.

\section{Method}
\label{sec:method}

In this section, we start with preliminary knowledge of centerline graph and Bézier curve, then introduce the architecture design and the proposed modules.

\subsection{Centerline Graph}
\label{sec:method:background}

\begin{figure*}[ht]
    \centering
    \includegraphics[width=0.99\linewidth]{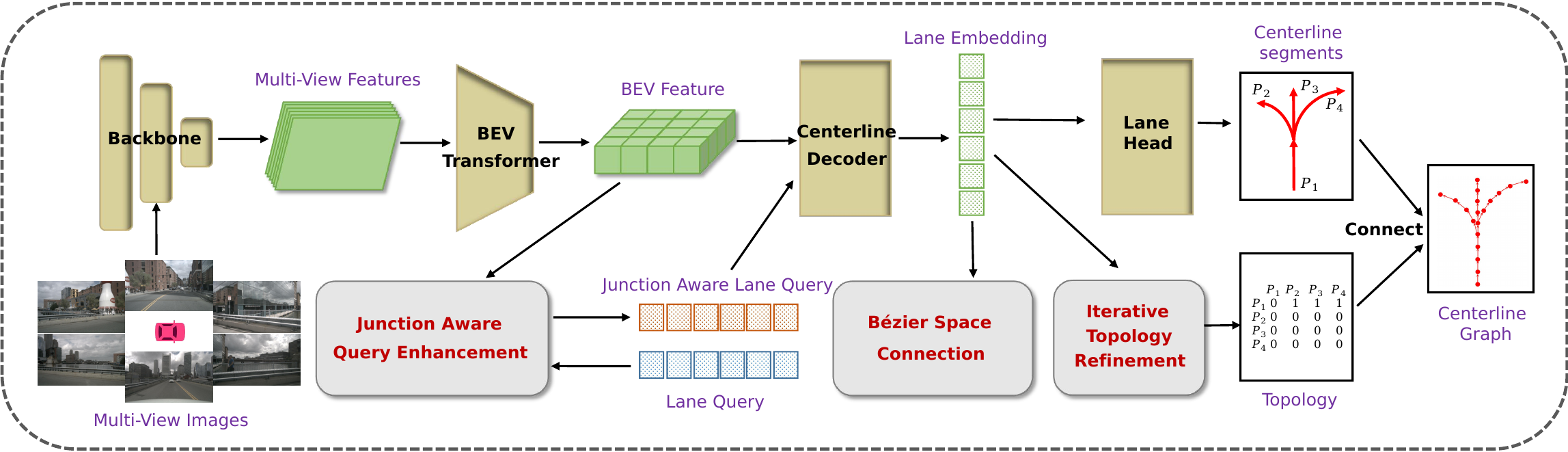}
    \caption{The overview architecture of CGNet. CGNet following the end-to-end paradigm of DETR, which takes 6 surrounding view images as inputs and output centerline graph without any post-processing. Three elaborately designed modules are utilized to preserving continuity, helping predict a continuous and smooth path. }
    \label{fig:overview}
\end{figure*}

Centerline graph refers to the topological structure that converts the centerlines of the lanes into a series of non-overlapped segments and connectivity among them, which is often used to describe the shape and connection relationship of the drivable path\cite{can2021structured,can2022topology,can2023improving,li2023topology}. Specifically, the graph is defined as $G=(V,E)$, where the vertices $V=\{P_{0},...P_{T-1}\}$ represents the set of centerline segments and the edges $E\subseteq \{(x,y)|(x,y)\in V^{2}\}$ represents the connectivity among segments. Each segment is represented as a set of ordered points denoted by $P=[p_{0},...p_{n-1}]$, where each point $p=(x,y) \in \mathbb{R}^{2}$. In this work, the edges(i.e. connectivity) are represented by an adjacency matrix $\mathcal{A}$ of the graph $G$, and $\mathcal{A}$ is directly outputted by the network along with the centerline segments.

\subsection{Preliminary on Bézier Curve.}
\label{sec:preliminarybc}
Bézier curve is a type of parametric curve that is widely used in computer graphics and path planning. It offers smooth and continuous curve in a compact and consistent form, which is defined by $m$ control points:
\begin{align}
    P(t)&=\sum_{i=0}^{m-1}b_{i}(t)c_{i}, 0\leq t \le 1
    \label{eq:Bézier1}
\end{align}
where, $c_{i}$ is the $i-th$ control point, $m$ is the degree of the curve and $b_{i}$ are Bernstein basis polynomials of degree $m$, which is formulated as:
\begin{align}
    b_{i}(t)&=\tbinom{m-1}{i}t^{i}(1-t)^{m-1-i},i=0,...,{m\!-\!1}.
    \label{eq:Bézier2}
\end{align}
In our work, a ground truth centerline is annotated by a set of 2D ordered points $P$ of polyline in the BEV coordinate frame. According to the Eq.\ref{eq:Bézier1}, we can obtain the corresponding Bézier control points $\mathcal{C}=[c_0,...,c_{m-1}]$ from polyline by using standard least squares fitting:
\begin{align}
\setlength{\arraycolsep}{2.5pt}
    \begin{bmatrix}  
      c_{0}\\  
      c_{1}\\ 
      \vdots \\  
      c_{m-1} \\ 
    \end{bmatrix}=
    \begin{bmatrix}  
      b_{0}(t_0)&\cdots&b_{m-1}(t_0) \\  
      b_{0}(t_1)&\cdots&b_{m-1}(t_1) \\ 
      \vdots&\ddots&\vdots \\  
      b_{0}(t_{n-1})&\cdots&b_{m-1}(t_{n-1}) \\ 
    \end{bmatrix}^{-1}
    \begin{bmatrix}  
      p_{0}\\  
      p_{1}\\ 
      \vdots \\  
      p_{n-1} \\ 
    \end{bmatrix}
    \label{eq:Bézier3}
\end{align}
$\{t_{i}\}_{i=0}^{n-1}$ is uniformly sampled from 0 to 1. For simplicity, Eq.\ref{eq:Bézier3} can be formulated as :
\begin{align}
    \mathcal{C}&=\mathcal{B}P
    \label{eq:Bézier4}
\end{align}
where $\mathcal{C} \in \mathbb{R}^{m\times 2}$ is the vector of all control points, $P \in \mathbb{R}^{n\times 2}$ is the vector of all polyline points. $\mathcal{B} \in \mathbb{R}^{m\times n}$ is the pseudo-inverse of Bernstein matrix, which conducts conversion between this two kind of points.

\subsection{CGNet}

\begin{figure*}[ht]
    \centering
    \includegraphics[width=0.99\linewidth]{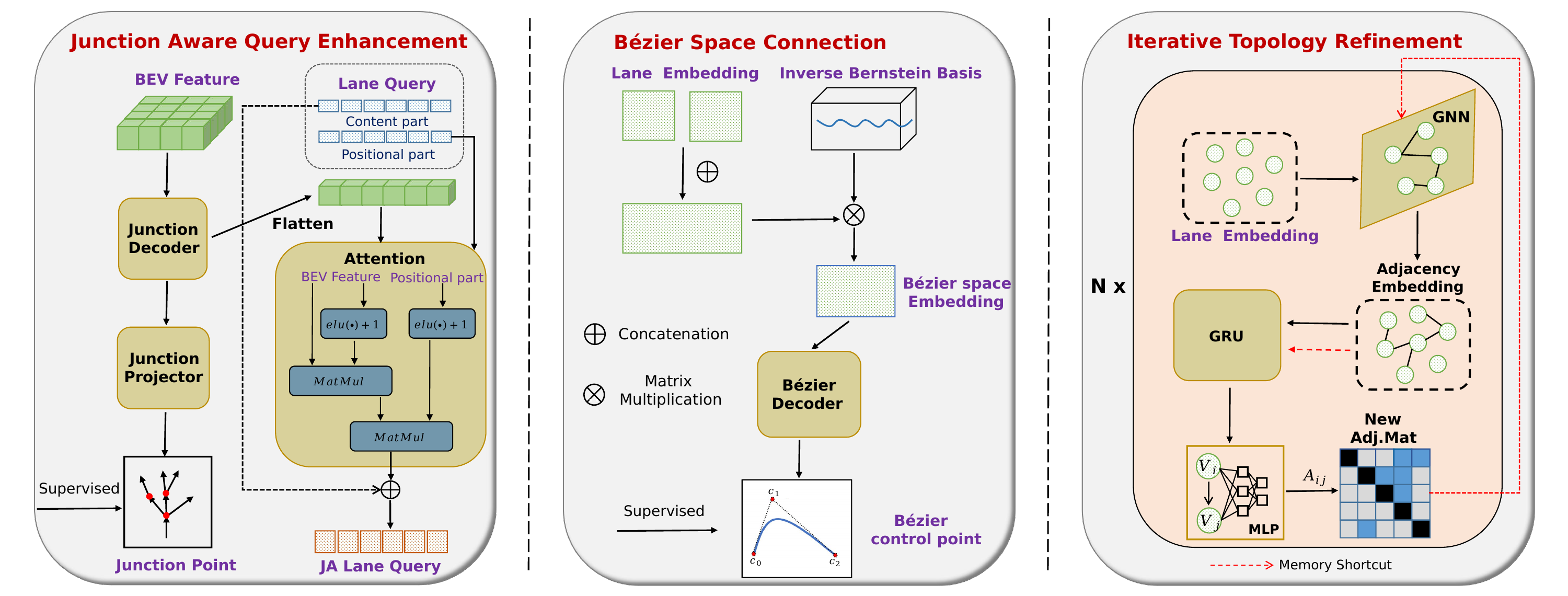}
    \caption{The detailed illustration of our proposed modules. Left: \emph{Junction Aware Query Enhancement Module}, which provides positional prior of junction to queries. Middle: \emph{Bézier Space Connection Module}, which enforces continuity constraints on any two connected segments in the Bézier space. Right: \emph{Iterative Topology Refinement Module}, which outputs connectivity of centerline segments in an iterative manner. "Memory Shortcut" means taking the output from previous layer.}
    \label{fig:module}
\end{figure*}

\subsubsection{Architecture Overview.}
\label{sec:overview}
The overall architecture of CGNet is illustrated in Fig.\ref{fig:overview}, which follows the end-to-end paradigm of DETR\cite{carion2020end}. Multi-view images from onboard cameras are used to generate features using a shared backbone. These features are then transformed into BEV space $\mathcal{F}^{bev}$ via a BEV encoder. The proposed \emph{Junction Aware Query Enhancement Module} takes a set of learnable lane queries $\{Q_i\}_{i=0}^{N-1}$ and BEV feature to output a set of enhanced junction aware queries $\{\tilde{Q}_i\}_{i=0}^{N-1}$, which capture positional information of junction points. Then queries interact with BEV feature through a multi-layer centerline decoder, which is a conventional transformer-based decoder, to obtain lane segment embeddings $\{\mathcal{E}_i\}_{i=0}^{N-1}$. Then \emph{Bézier Space Connection Module} projects embeddings of any two logically connected segments into a Bézier space, predicts control points in a compact form, and use the $L_1$ loss to supervise control points. In this way, the segment embeddings are constrained to be spatially continuous in feature space. Furthermore, the embeddings of any two connected segments exhibit greater similarity compared to those disconnected pairs. Based on these lane embeddings, the proposed \emph{Iterative Topology Refinement Module} identifies centerline segments that are similar to each other at the feature-level and establishes corresponding connectivity. The lane embeddings are also taken by a MLP-based Lane Head to predict centerline segments.

\subsubsection{Junction Aware Query Enhancement Module.}
\label{sec:jaq}
Since we adopt a DETR like framework, the design of lane query is important for this problem. Inspired by recent object detection methods \cite{meng2021conditional,liu2022dab} like DAB-DETR, in which an object query consists of two parts: a content part and a positional part. Box coordinates $(x,y,h,w)$ are proper positional prior for object queries. Experiments demonstrate effectiveness of this kind of positional prior. However, unlike object detection, where positional prior can be well represented by a box, we detect centerline segments which is represented by a set of ordered points. It is challenge to directly provide positional prior of an elongated segment for lane queries. Since junction points are critical for both detecting lane segments and recovering topology, we design a sub-network \emph{Junction Aware Query Enhancement Module}, \emph{JAQ}, to encode junction points as positional prior.

As shown in Fig.\ref{fig:module} left, \emph{JAQ} consists of two branches. Specifically, the first branch employs a Junction Decoder to decode the BEV feature into the junction feature, which can be formulated as:
\begin{align}
    \mathcal{F}^{jp}&=\Theta(\mathcal{F}^{bev})   
    \label{eq:jd}
\end{align}
where $\mathcal{F}^{jp}$ is the junction feature, $\Theta$ is the junction decoder, which consists of several convolution and BN\cite{ioffe2015batch} layers. The junction projector, which has a similar structure of the junction decoder, takes the junction feature $\mathcal{F}^{jp}$ to generate a heatmap of junction points. Then, we introduce the junction aware loss $\mathcal{L}_{ja}$ to supervise learning of the heatmap, which is formulated as follows:
\begin{align}
    \mathcal{L}_{ja}&=\mathcal{L}_{focal}(Z, Z^{gt})
    \label{eq:ljd}
\end{align}
where $Z\in \mathbb{R}^{H_{bev}\times W_{bev}}$ is the predicted heatmap, $Z^{gt} \in \mathbb{R}^{H_{bev}\times W_{bev}}$ is the ground truth junction points map. Due to the sparse distribution of junction points across the entire map, we dilate the junction points by a radius $R$ to reduce the difficulty of learning. $\mathcal{L}_{focal}$ is the modified focal loss for keypoint detection\cite{law2018cornernet}. In this way, we obtain a group of features, which is termed as junction feature. The second branch splits the input lane queries $Q$ into a content part $Q^{c}$ and a positional part $Q^{p}$. The positional part $Q^{p}$ interacts with the junction feature $\mathcal{F}^{jp}$ to obtain positional prior of junction points through an attention layer, which is formulated as:
\begin{align}
    \tilde{Q}^{p}&=Attention(Q^{p},\mathcal{F}^{jp},\mathcal{F}^{jp})
    \label{eq:la}
\end{align}
where $\tilde{Q}^{p}$ is the enhanced positional part of a query. To save computation, we adopt an efficient linear attention\cite{katharopoulos2020transformers} as the attention layer. Finally, we concatenate content part $Q^{c}$ and positional part $\tilde{Q}^{p}$ to obtain a junction aware lane query $\tilde{Q}$. The \emph{JAQ} is able to enhance continuity from a point-level perspective and our ablation studies demonstrate effectiveness of this module.

\subsubsection{Bézier Space Connection Module.}
\label{sec:bsc}
To further enhance continuity from a segment-level perspective, we propose the \emph{Bézier Space Connection Module(BSC)}. This module represents any two connected centerline segments by a unique set of Bézier control points. This means that the centerlines of these connections are treated as a whole and represented in a more compact and consistent form, thereby ensuring better continuity and smoothness. Previous works\cite{feng2022rethinking, qiao2023end} have introduced Bézier curve for lane detection, which directly predict the control points from the network and then convert control points into polyline points. This conversion occurs in a two-dimensional Cartesian coordinates space. Different from these works, we utilize the Bézier curve in a novel way, where the conversion takes place in a high-dimensional feature space, i.e. the lane embedding space. Fig.\ref{fig:module} middle illustrates our method. Specifically, we denote the set of any two logically connected lane embeddings as $\mathbb{S}$, which are obtained by Hungarian matching (Sec.\ref{sec:e2el}) according to ground truth. Then we concatenate lane embeddings and project them into the Bézier space:
\begin{align}
    \hat{\mathcal{E}}&=\mathcal{B} \, (\mathcal{E}_i \oplus \mathcal{E}_j), \quad (\mathcal{E}_i, \mathcal{E}_j) \in \mathbb{S}
    \label{eq:Bézier5}
\end{align}
where $\mathcal{E} \in \mathbb{R}^{n\times C}$ is the lane embedding, $n$ denotes the number of points on a polyline. $\hat{\mathcal{E}} \in \mathbb{R}^{m\times C}$ is a new lane embedding in Bézier space, $m$ denotes the number of control points of Bézier curve. $\mathcal{B} \in \mathbb{R}^{m\times 2n}$ is the pre-calculated pseudo-inverse of Bernstein matrix. Finally, the Bézier space embedding $\hat{\mathcal{E}}$ is fed into a Bézier decoder consisting of a few MLP layers, to output the control points $\mathcal{C}$. The ground truth control points, denoted as $\mathcal{C}^{gt}$, are obtained by connecting any two centerline segments according to the ground truth adjacent matrix and calculated using Eq.\ref{eq:Bézier3}. Here, we leverage the L1-loss to supervise the learning, which can be formulated as:
\begin{align}
    \mathcal{L}_{bezier}&=\frac{1}{|\mathcal{C}^{gt}|}\sum_{i=1}^{|\mathcal{C}^{gt}|}||\mathcal{C}_i-\mathcal{C}_{i}^{gt}||_{1}
    \label{eq:Bézier6}
\end{align}

Unlike traditional Bézier-based lane detection works, where Bézier constraint is added on the final output, we fuse embeddings of logically connected segments and project them into Bézier space to capture knowledge of consistency and smoothness at an earlier stage. The lane embedding implicitly carries information about continuity in a more global perspective, which enhances the accuracy of positional prediction. Moreover, the embeddings of any two connected centerline segments exhibit greater similarity compared to those disconnected pairs, thereby aiding the following topology prediction. Our ablation studies on this early or post fusion strategy demonstrate effectiveness of this idea.

\subsubsection{Iterative Topology Refinement Module.}
\label{sec:itr}
The connectivity prediction task is inherently defined on a graph, which has complex and entangled structural information. Thus, we propose a graph-based module, the \emph{Iterative Topology Refinement Module(ITR)}, which consumes lane embeddings to produce adjacency embeddings for connectivity prediction. The approach that utilizing prediction of previous layer to improve  the prediction of current layer in the multi-layer transforemr decoder has been proved to be effective in object detection task\cite{liu2022dab}. We further design this module in an iterative refinement manner.

Fig.\ref{fig:module} right shows the details of \emph{ITR} module. Specifically, we first input lane embeddings $\mathcal{E}$ and adjacency matrix $\mathcal{\hat{A}}$ into the GCN\cite{kipf2016semi} to produce adjacency embeddings $\Lambda$, which can be formulate as:
\begin{align}
    \Lambda^{l}&=GCN(\mathcal{E}^{l}, \mathcal{\hat{A}}^{l-1})
    \label{eq:gcn}
\end{align}
where $l$ is the $l$-th layer, the adjacency matrix $\mathcal{\hat{A}}^{l-1}$ is the output from previous layer. Note that the initial adjacency matrix is set to all zeros, which means no connection among lane embeddings. Then we leverage GRU\cite{chung2014empirical} to process the adjacency embeddings $\Lambda$ along with the $\Lambda^{l-1}$ from previous layer, the output from GRU is taken by a MLP to predict adjacency matrix:
\begin{align}
    \mathcal{\hat{A}}^{l}&=MLP(GRU(\Lambda^{l}, \Lambda^{l-1}))
    \label{eq:gru}
\end{align}
The initial adjacency embeddings are also set to all zeros. The GCN and GRU of all layers share the same parameters. Since each layer of the decoder is supervised by the ground truth, the prediction can be iteratively refined.

\subsection{End-to-end Training}
\label{sec:e2el}

\textbf{Prediction and Label Matching.} CGNet infers a fixed-size set of $N$ predictions and assuming $N$ is larger than the number of centerline segments. To achieve end-to-end learning, we leverage Hungarian algorithm to find a suitable label assignment $\hat{\sigma}$ between labels and predictions:
\begin{align}
    \hat{\sigma}&=\underset{\sigma \in \Pi_{N}}{\arg\min}\sum_{i}^{N} \mathcal{L}_{Hungarian}(y_i, \hat{y}_{\sigma(i)})
    \label{eq:optimize}
\end{align}
where $y_i$ is the ground truth label, $\hat{y}_{\sigma}$ is one of the permutation of $N$ predictions $\Pi_{N}$. The matching cost of Hungarian algorithm considers both the class label and the position:
\begin{align}
    \mathcal{L}_{Hungarian}&= \mathcal{L}_{c}(\hat{l}_{\sigma(i)}, l_i) + \mathcal{L}_{p}(\hat{P}_{\sigma(i)}, P_i)
    \label{eq:hungarian}
\end{align}
where $\mathcal{L}_{c}$ is the focal loss\cite{lin2017focal} that calculates the cost between the predicted classification score $\hat{l}_{\sigma(i)}$ and GT class label $l_i$. $\mathcal{L}_{p}$ is the L1-loss that calculates the distance between predicted polyline $\hat{P}_{\sigma(i)}$ and GT polyline $P_i$. Note that, there is only one class in our problem setting, so the classification score reflects the probability of 'no lane' $\varnothing$. 

\textbf{Training Loss.} Based on the matching results, we adopt classification loss, ployline loss and topology loss for training. The classification loss is a Focal Loss formulated as:
\begin{align}
    \mathcal{L}_{cls}&=\sum_{i=0}^{N-1}\mathcal{L}_{Focal}(\hat{l}_{\sigma(i)}, l_i)
    \label{eq:cls}
\end{align}
The polyline loss is a L1-loss, which calculates the point to point distance between predicted polyline and GT polyline:
\begin{align}
    \mathcal{L}_{poly}&=\sum_{i=0}^{N-1} \mathds{1}_{\{c_i \neq \varnothing \}} \sum_{j=0}^{n-1}||\hat{p}_{\sigma(i), j}-p_{i, j}||_1
    \label{eq:pos}
\end{align}
Following STSU\cite{can2021structured}, we employ the binary cross-entropy loss as topology loss, which is formulated as:
\begin{align}
    \mathcal{L}_{topo}&=\frac{1}{T^2}\sum_{i=0}^{T}\sum_{j=0}^{T}\mathcal{L}_{bce}(\hat{a}_{i,j}, a_{i,j})
    \label{eq:topo}
\end{align}

where $a_{i,j} \in \{0,1\}$ is the $ij-th$ element in GT adjacent matrix $\mathcal{A} \in \mathbb{R}^{T\times T}$, $T$ is the number of GT centerline segments. $\hat{a}_{i,j} \in [0,1]$ is the predicted probability of connectivity. In addition, the Direction Loss\cite{liao2022maptr} $L_{dir}$ is employed to restrict the geometrical shape. With the previously introduced $L_{ja}$ and $L_{bézier}$, the overall loss is a weighted sum of all losses:
\begin{align}
    \begin{split}
    \mathcal{L}&=\lambda_{1} \mathcal{L}_{cls} +\lambda_{2} \mathcal{L}_{poly} +\lambda_{3} \mathcal{L}_{topo} \\
    & +\lambda_{4} \mathcal{L}_{dir} +\lambda_{5} \mathcal{L}_{bezier} +\lambda_{6} \mathcal{L}_{ja}
    \end{split}
\end{align}
where $\lambda_i$ is weighting factor.

\begin{figure*}[ht]
    \centering
    \includegraphics[width=0.90\linewidth]{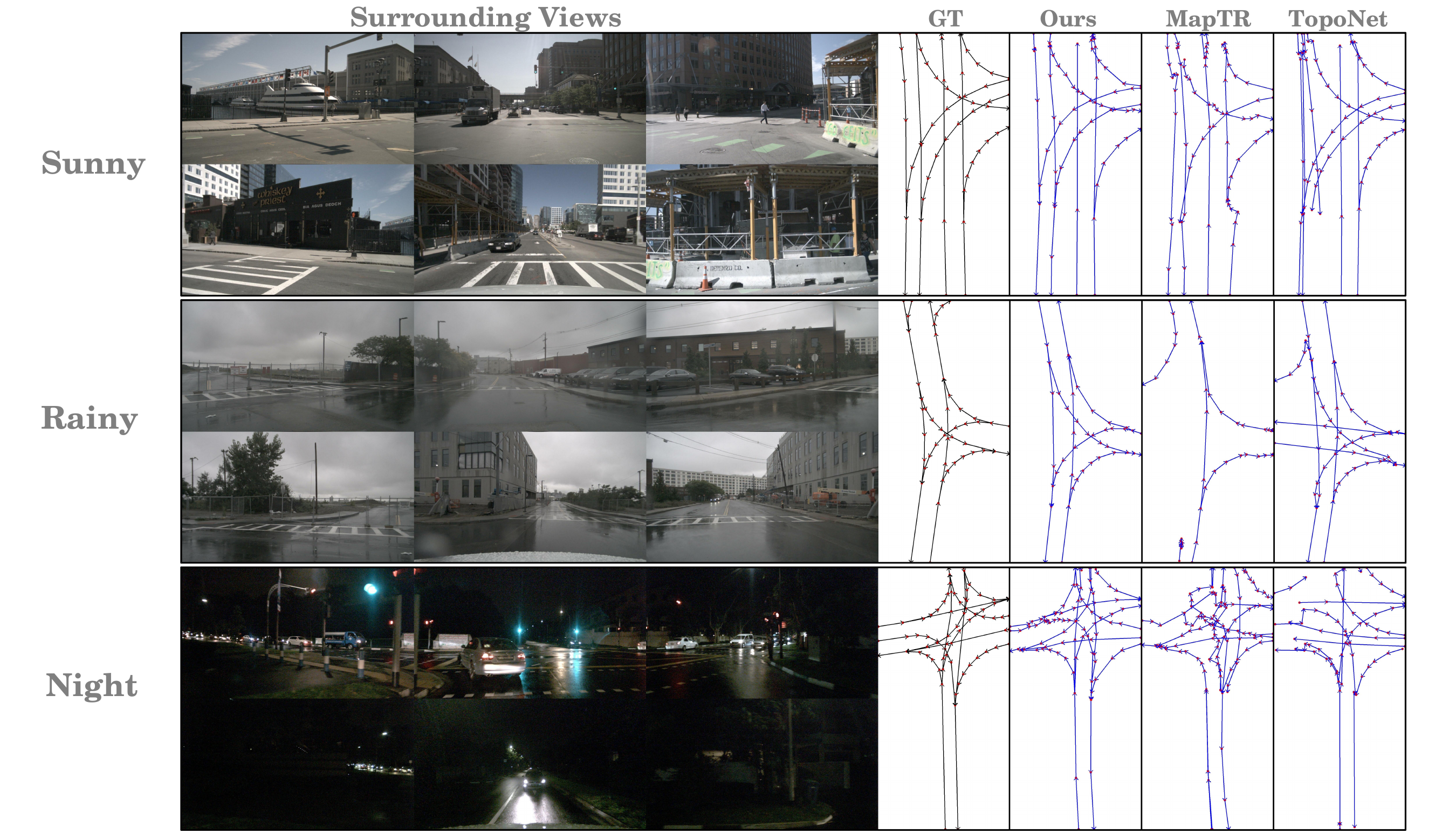}
    \caption{Qualitative comparisons under different weather and lighting conditions on nuScenes. CGNet predicts more accurate position of junction points and correct topology, leading to a more continuous and smooth path compared to MapTR and TopoNet. CGNet demonstrats stronger robustness under different conditions.}
    \label{fig:result}
\end{figure*}

\section{Experiments}
\label{sec:exp}

\begin{table*}[ht]
    \centering
    \caption{Comparison on nuScenes using point-level metrics. The best results are marked in {\color{red}{red}} and the second best are in {\color{blue}{blue}}, `-' indicates no results, `$\ast$' indicates different perception range setting and `$\dagger$' indicates the method is re-implemented by us. FPSs are measured on the same machine with single RTX 3090. CGNet achieves the second best in FPS, which can also realize real-time centerline graph construction.}
    \resizebox*{0.98 \linewidth}{!}{
        \begin{tabular}
            {
                >{\arraybackslash}p{2.9cm}
                >{\arraybackslash}p{1.4cm}|
                >{\arraybackslash}p{2.0cm}
                >{\arraybackslash}p{2.0cm}
                >{\arraybackslash}p{2.0cm}
                >{\arraybackslash}p{1.6cm}
                >{\arraybackslash}p{1.6cm}|
                >{\arraybackslash}p{1.0cm}
                >{\arraybackslash}p{1.2cm}
            }
            \hline
		{Methods} &{Epoch} &{GEO F1$\uparrow$} &{TOPO F1$\uparrow$} &{JTOPO F1$\uparrow$} &{APLS$\uparrow$} &{SDA$\uparrow$} &{FPS$\uparrow$} &{Params.$\downarrow$}\\
            \hline
            $\ast$STSU\cite{can2021structured}              &200 &33.0 &20.6 &13.9 &11.0 &7.0 &- &-\\
            $\dagger$HDMapNet\cite{li2022hdmapnet}          &30 &45.5 &20.0 &14.8 &25.9 &0.5 &- &-\\
            $\dagger$VectorMapNet\cite{liu2023vectormapnet} &24 &48.4 &38.1 &27.9 &10.3 &7.2 &6.1 &\textcolor{blue}{36.9M}\\
            $\dagger$TopoNet\cite{li2023topology}           &24 &50.8 &39.6 &31.9 &26.2 &5.6 &10.1 &40.3M\\
            $\dagger$MapTR\cite{liao2022maptr}              &24 &53.3 &39.7 &32.4 &26.8 &7.6 &12.7 &36.0M\\ 
            CGNet(Ours)                                     &24 &54.7 &42.2 &34.1 &30.7 &8.8 &11.2 &38.8M\\
            \hline
            $\dagger$TopoNet\cite{li2023topology} 
            &110 &59.9 &49.6 &39.5 &37.4 &11.6 &10.1 &40.3M\\
            $\dagger$MapTR\cite{liao2022maptr}    
            &110 &\textcolor{blue}{61.2} &\textcolor{blue}{50.0} &\textcolor{blue}{40.7} &\textcolor{blue}{38.4} &\textcolor{blue}{13.4} &\textcolor{red}{12.7} &\textcolor{red}{36.0M}\\
            CGNet(Ours)                                     
            &110 &\textcolor{red}{63.9} &\textcolor{red}{53.2} &\textcolor{red}{43.3} &\textcolor{red}{41.4} &\textcolor{red}{14.5} &\textcolor{blue}{11.2} &38.8M\\
            \hline
        \end{tabular}
        }
    \label{table:PL metrics}
\end{table*}

\subsection{Dataset}

We evaluate CGNet on the challenging nuScenes\cite{caesar2020nuscenes} and Argoverse2\cite{wilson2021argoverse} datasets. NuScenes contains 1000 scenes of roughly 20s duration each and key samples are annotated at 2Hz. Argoverse2 contains 1000 logs and each log provides 15s of 20Hz RGB images. Both nuScenes and Argoverse2 directly provides the centerline segments and the connectivity among these segments.

\subsection{Evaluation Metrics}
To evaluate the quality of CGNet, especially in terms of measuring continuity, we adopt fine-grained point-level metrics (Geo F1, TOPO F1, JTOPO F1, APLS, SDA). These metrics are mainly borrowed from previous works\cite{buchner2023learning,he2022lane,van2018spacenet}, which focus on extracting lane graph from aerial imagery. We modify some parameters to meet the accuracy requirements of centerline graph. To achieve a more comprehensive comparison, we also use segment-level metrics (IoU, mAP$_{cf}$, DET$_l$, TOP$_{ll}$) for evaluation, as done in previous works\cite{can2021structured,li2023topology,liao2022maptr,liu2023vectormapnet,wang2023openlane}.

In order to use point-level metrics, we generate a point-level graph $\ddot{G}\!=\!\!(\ddot{V},\ddot{E})$ from a segment-level graph(introduced in Sec.\ref{sec:method:background}), where $\ddot{V}$ represents all points of all polylines and $\ddot{E}$ represents all edges, i.e. the connectivity among these points. In the following section, we provide a brief introduction to point-level metrics. Please refer to our supplementary material for more details. 

\textbf{GEO metric.} This metric\cite{he2022lane} computes a one-to-one matching of vertices between prediction $\hat{\ddot{G}}$ and GT $\ddot{G}$. The F1-score is utilized to evaluate the prediction accuracy according to the matching result.

\textbf{TOPO / JTOPO metrics.} GEO metric focuses on local correctness, but it does not take connectivity into account. TOPO metric\cite{he2022lane} constructs a sub-graph for each matched vertex pair, then computes GEO metric between two sub-graphs. JTOPO metric\cite{liao2023lane} highlights the correct connection of the junction points based on the TOPO metric.

\textbf{APLS metric.} This metric\cite{van2018spacenet} cares about both logical topology as well as the physical topology of the lane. It sums the differences in optimal path lengths between vertices in the predicted graph and GT graph. 

\textbf{SDA metric.} This metric\cite{buchner2023learning} evaluates the precision of predicted junction points within a circular area of radius $R$ from given ground truth junction points.

\subsection{Implementation Details}
\label{sec:imp}

Following previous methods\cite{liao2022maptr,liu2023vectormapnet,ding2023pivotnet,qiao2023end}, we set the perception range to [-15.0m, 15.0m] along the $X$-axis and [-30.0m, 30.0m] along the $Y$-axis. We adopt ResNet50\cite{he2016deep} with FPN\cite{lin2017feature} as backbone and GKT\cite{chen2022efficient} to obtain BEV features. In the centerline decoder, we adopt Deformable Attention\cite{zhu2020deformable} to make queries interact with BEV features. We set the size of each BEV grid to 0.3m, the number of lane query $N$=50 and the number of points of a polyline $n$=20. Our model is trained and evaluated using PyTorch, on Tesla A100 GPUs with batch size 32. The AdamW optimizer and cosine annealing schedule is employed with a learning rate of $6e^{-4}$. The weighting factors$\{\lambda_1,\lambda_2,\lambda_3,\lambda_4,\lambda_5,\lambda_6\}$ are set to $\{2,5,1,0.005,0.01,0.1\}$ respectively.

\subsection{Comparison with State-of-the-Arts}
\label{sec:expsota}

\textbf{Test on point-level metrics.} Since only a few methods have been proposed for online centerline graph construction using surround-view cameras, we carefully modify some algorithms to suit centerline graph learning task. STSU\cite{can2021structured} only utilize front-view images as input, so we reset the perception range to [-15.0m, 15.0m, 0.0m, 30m]. VectormapNet\cite{liu2023vectormapnet} and MapTR\cite{liao2022maptr} are not designed for topology modeling, so we directly borrow code from STSU for connectivity estimation. In addition, MapTR ignores the direction with the permutation-equivalent modeling, we abandon this design to learn a consistent direction. HDMapNet\cite{li2022hdmapnet} is built upon image segmentation, which is difficult to directly incorporate topology learning module into the network, so we obtain connectivity based on the distance of points in post-processing to build the centerline graph. TopoNet\cite{li2023topology} focuses on the driving scene topology, including the topology of centerline segments and traffic elements. Thus, we carefully remove the traffic elements part and only retain the centerline part. To achieve a fair comparison, our modifications follow the principle of making minimal changes. More implementation details can be found in our supplementary material. 

We compare CGNet with state-of-the-art methods on nuScenes in Table \ref{table:PL metrics}. CGNet outperforms all other methods by a significant margin across all point-level metrics on the nuScenes validation set. CGNet outperforms the second-best methods by 4.4\% in GEO F1 and 6.4\% in TOPO F1, which demonstrates its ability on improving accuracy on both point detection and topology prediction. Additionally, the JTOPO F1 and SDA are improved by 6.4\% and 8.2\%, which indicate that junction points of CGNet are more accurate. Furthermore, the APLS  are improved by 7.8\%, which indicates that CGNet is able to construct a more continuous path. We also conduct experiments on Argoverse2 in Table \ref{table:av2}, which demonstrates the generalization ability of our scheme.

\begin{table}[ht]
    \centering
    \caption{Results on Argoverse2 val.}
    \renewcommand{\arraystretch}{1.0}
    \resizebox*{0.6 \linewidth}{!}{
        \begin{tabular}
            {
                >{\arraybackslash}p{1.2cm} 
                >{\centering\arraybackslash}p{1.2cm}|
                >{\centering\arraybackslash}p{1.6cm}
                >{\centering\arraybackslash}p{2.0cm}
                >{\centering\arraybackslash}p{1.2cm}
                >{\centering\arraybackslash}p{1.2cm}
                >{\centering\arraybackslash}p{1.2cm}
            }
            \hline
		{Methods} &{Epoch} &{TOPO F1$\uparrow$} &{JTOPO F1$\uparrow$} &{APLS$\uparrow$} &{SDA$\uparrow$} &{TOP$_{ll}$ $\uparrow$}\\
            \hline
            TopoNet     
            &6 &30.2 & 23.7 &15.3 &7.7 &0.3 \\
            MapTR    
            &6 &\textcolor{blue}{42.8} &\textcolor{blue}{33.5} &\textcolor{blue}{22.3} &\textcolor{blue}{13.6} &\textcolor{blue}{0.5}\\
            Ours                                    
            &6 &\textcolor{red}{44.5} &\textcolor{red}{34.6} &\textcolor{red}{23.6} &\textcolor{red}{13.7} &\textcolor{red}{0.5}\\
            \hline
        \end{tabular}
        }
    \label{table:av2}
\end{table}

\begin{table}[ht]
    \centering
    \caption{Comparison on nuscenes using segment-level metrics. The best results are marked in {\color{red}{red}} and the second best are in {\color{blue}{blue}}.}
    \renewcommand{\arraystretch}{1.1}
    \resizebox*{0.4 \linewidth}{!}{
        \begin{tabular}{ccccc}
            \hline
		{Methods} &{IoU} &{mAP$_{cf}$} &{DET$_{l}$} &{TOP$_{ll}$}\\
            \hline
            VectorMapNet\cite{liu2023vectormapnet} &29.8 &25.5 &16.1 &0.5  \\
            TopoNet\cite{li2023topology}             &51.3 &31.6 &17.9 &0.8   \\
            MapTR\cite{liao2022maptr}              &\textcolor{blue}{53.7} &\textcolor{blue}{33.1} &\textcolor{blue}{18.9} &\textcolor{blue}{1.0}  \\
            \hline
            CGNet(Ours) &\textcolor{red}{56.3} &\textcolor{red}{35.2} &\textcolor{red}{22.0} &\textcolor{red}{1.3}\\
            \hline
        \end{tabular}
        }
    \label{table:SL metrics}
\end{table}

\textbf{Test on segment-level metrics.} To achieve a more comprehensive comparison, we also directly evaluate the centerline segments using segment-level metrics. As shown in Table \ref{table:SL metrics}, CGNet surpasses all the methods on these segment-level metrics. In summary, evaluation results from point-level and segment-level all validate the effectiveness of CGNet.
Fig.\ref{fig:result} provides some qualitative comparisons with MapTR and TopoNet under different weather and lighting conditions.

\subsection{Ablation Analysis}
\label{sec:ablation}

In this section, we first conduct a series of ablation studies on nuScenes to evaluate the effectiveness of the proposed modules in Table \ref{table:single module}. Then we analyze the impact of different parameter and architecture choices on CGNet. Because there are multiple combinations of these factors, we only test a specific component of our approach in isolation. Table \ref{table:ablation} summarizes these results in great details. All ablation experiments are trained with 24 epochs.

\begin{table}[ht]
    \centering
    \caption{Effectiveness of different modules in CGNet.}
    \resizebox*{0.6 \linewidth}{!}{
        \begin{tabular}{ccc|cccc|cc}
            \hline
		{JAQ} &{BSC} &{ITR} &{TOPO F1$\uparrow$} &{JTOPO F1$\uparrow$} &{APLS$\uparrow$} &{SDA$\uparrow$} &{FLOPs} &{Params.}\\
            \hline
                \usym{2717} &\usym{2717} &\usym{2717} &39.7 &32.4 &26.8 &7.6 &216.7G &36.0M\\
                \usym{2713} &\usym{2717} &\usym{2717} &40.5 &33.1 &28.5 &7.7
                &241.6G &37.8M\\
                \usym{2717} &\usym{2713} &\usym{2717} &41.4 &33.4 &29.1 &8.3
                &216.7G &36.0M\\
                \usym{2717} &\usym{2717} &\usym{2713} &40.1 &32.7 &28.6 &7.8
                &217.0G &36.9M\\
                \usym{2713} &\usym{2713} &\usym{2713} &\textbf{{42.2}} &\textbf{{34.1}} &\textbf{{30.7}} &\textbf{{8.8}} &241.9G &38.8M\\
            \hline
        \end{tabular}
        }
    \label{table:single module}
\end{table}

\textbf{Effectiveness of different modules.} Table \ref{table:single module} shows detailed ablation results of each module and lists the computational overhead. Specifically, the first row represents the baseline method, which is our modified MapTR (mentioned in \ref{sec:expsota}) for centerline graph learning. The following three row shows the results when adopting the \emph{JAQ} module, \emph{BSC} module and \emph{ITR} module separately. All modules lead to performance improvements across all metrics. Note that, the \emph{BSC} module is only used during the training phase and therefore does not introduce additional computational overhead.

\textbf{Ablation for Bézier Control Points.} Bézier curve can be represented by different numbers of control points and more control points can represent more complex curve. We empirically analyze the corresponding performance gain by changing the number of control points. As shown is Tabel \ref{table:ablation} first row, when more control points are used, the performance is gradually decreased. This is because when the number of control points increases, the difficulty of neural network learning also increases. Moreover, the shape of the centerline in real scenarios is usually simple (e.g. straight lines), so using a small number of control points is sufficient.

\textbf{Ablation for Dilated Radius.} As mentioned in Sec.\ref{sec:jaq}, we dilate the ground truth junction points heatmap by a radius $R$. We evaluate the influence of different radius setting in Table \ref{table:ablation} second row. The best results is achieved in the middle setting $R$=9. This is because the smaller radius setting may suffer from data imbalance and the larger radius setting influences the positional accuracy.

\begin{table}[t]
\centering
\caption{Ablation analysis for different settings of our CGNet. \underline{Underline} indicates the default settings in our model.}
\resizebox*{0.6\linewidth}{!}{
    \begin{tabular}
        {
            >{\arraybackslash}p{2.0cm} 
            >{\centering\arraybackslash}p{1.5cm}|
            >{\centering\arraybackslash}p{1.8cm}
            >{\centering\arraybackslash}p{1.8cm}
            >{\centering\arraybackslash}p{1.0cm}
            >{\centering\arraybackslash}p{1.0cm}
        }
        \hline
        {Experiment}  & {Method}  &{TOPO F1$\uparrow$} &{JTOPO F1$\uparrow$} &{APLS$\uparrow$} &{SDA$\uparrow$} \\

        \hline
        \multirow{3}{*}{Nums. Pts.} &\underline{4} &42.2 &34.1 &30.7 &8.8 \\
        &6 &40.8 &33.2 &28.9 &8.6 \\
        &8 &40.4 &32.6 &28.2 &7.9 \\
        \hline

        \hline
        \multirow{3}{*}{Dilated Radius} 
        &7 &40.4 &32.7 &28.7 &8.9 \\
        &\underline{9} &42.2 &34.1 &30.7 &8.8 \\
        &11 &40.9 &33.4 &28.8 &8.6 \\
        \hline

        \hline
        \multirow{2}{*}{Projection} 
        &Lane               &40.1 &32.7 &28.5 &7.0 \\
        &\underline{Feature} &42.2 &34.1 &30.7 &8.8 \\
        \hline

        \hline
        \multirow{3}{*}{Iteration}
        &Iter.1 &39.6 &31.5 &27.6 &7.1\\
        &Iter.3 &41.3 &33.3 &29.9 &8.3\\
        &\underline{Iter.6} &42.2 &34.1 &30.7 &8.8\\
        \hline
        
    \end{tabular}
    }
    \label{table:ablation}
\end{table}

\textbf{Ablation for Projection.} As introduced in Sec.\ref{sec:bsc}, we leverage the pseudo-inverse of Bernstein matrix $\mathcal{B}$ to project the lane embeddings into bézier space for predicting the control points. A more intuitive and general approach is to directly convert the finally predicted polyline into control points. We name these two approaches as "projection on feature" and "projection on lane" separately. Table \ref{table:ablation} third row provides the comparisons between two approaches and "projection on feature" achieves better performance. In addition, the proposed \emph{BSC} module can be viewed as an auxiliary task and will be removed during the inference phase, which saves computations.

\textbf{Ablation for Iteration.} We set the number of decoder layers to 6, as done in most previous works. The outputs from each layer are supervised by ground truth in a parallel way. Thus, we utilize predicted adjacency matrix of previous layer to improve the prediction of current layer. Similar practice is commonly used in object detection to refine box. Tab.\ref{table:ablation} fourth row demonstrates \emph{ITR} can iteratively refine topology prediction.

\section{Conclusion}
\label{sec:conclusion}

In this work, we introduce CGNet to build centerline graph online. By considering junction points as positional prior in lane queries, utilizing a global smooth constraint in a novel Bézier space and iteratively predict adjacency matrix, CGNet effectively mitigates the discontinuity issue compared to existing approaches as well as improve the accuracy of topology and segment position. We believe CGNet can be further extended towards production-grade online lane topology understanding and facilitate future research.


%
%
\bibliographystyle{splncs04}
\bibliography{main}
\end{document}